\documentclass[10pt,twocolumn,letterpaper]{article}

\usepackage{cvpr}              
\usepackage{multirow}
\usepackage{algorithm}
\usepackage{algorithmic}
\usepackage{soul}
\definecolor{cvprblue}{rgb}{0.21,0.49,0.74}
\usepackage[pagebackref,breaklinks,colorlinks,allcolors=cvprblue]{hyperref}

\usepackage [dvipsnames] { xcolor }
\usepackage{bbding}

\def\paperID{11853} 
\def\confName{CVPR}
\def\confYear{2025}

\title{BAHOP: Similarity-based Basin Hopping for A fast hyper-parameter search in WSI classification}

\author{Jun Wang\\
City University of HONG KONG\\
{\tt\small jwang699-c@my.cityu.edu.hk}
\and
Yu Mao\\
MBZUAI\\
{\tt\small yu.mao@mbzuai.ac.ae}
\and
Yufei Cui\\
McGill University\\
{\tt\small yufei.cui@mail.mcgill.ca}
\and
Nan Guan\\
City University of HONG KONG\\
{\tt\small nanguan@cityu.edu.hk}
\and
Chun Jason Xue\\
MBZUAI\\
{\tt\small jason.xue@mbzuai.ac.ae}
}

\begin{document}
\maketitle
\begin{abstract}
Pre-processing whole slide images (WSIs) can impact classification performance. Our study shows that using fixed hyper-parameters for pre-processing out-of-domain WSIs can significantly degrade performance. Therefore, it is critical to search domain-specific hyper-parameters during inference. However, searching for an optimal parameter set is time-consuming. To overcome this, we propose BAHOP, a novel Similarity-based Basin Hopping optimization for fast parameter tuning to enhance inference performance on out-of-domain data. The proposed BAHOP achieves 5\% to 30\% improvement in accuracy with $\times5$ times faster on average. 
\end{abstract}    
\section{Introduction}
\label{sec:intro}

Following the success of early computational pathology applications, dataset sizes have increased, prompting more multicentric efforts to address variability in staining, image quality, scanning characteristics, and tissue preparation in different laboratories. This has highlighted a known issue where computational pathology algorithms perform best on data on which they were trained but less well on data from other sources. Generalization continues to pose a major challenge in this field, as significant differences in clinical variables between tissue source sites can adversely affect the performance of histopathological tasks~\cite{van2021deep, tcga-biased, howard2021impact}.

State-of-the-art multiple instance learning (MIL) models have shown promising improvements in whole slide image (WSI) classification on out-of-domain (OOD) data by training on large, diverse datasets~\cite{zhang2022dtfd,rnnmil,lu2021data,transmil,dsmil,mhim-mil}. Out-of-domain classification occurs when a model is trained on one dataset and tested on another from a different domain. For instance, a model might be trained using Camelyon16 data~\cite{c17} and then evaluated using data from various Camelyon17 centers.
As indicated in Table~\ref{tab:important}, the performance of a model trained in Camelyon16 can vary significantly at specific centers within Camelyon17.

\begin{figure}[t]
    \centering
    \includegraphics[width=0.45\textwidth]{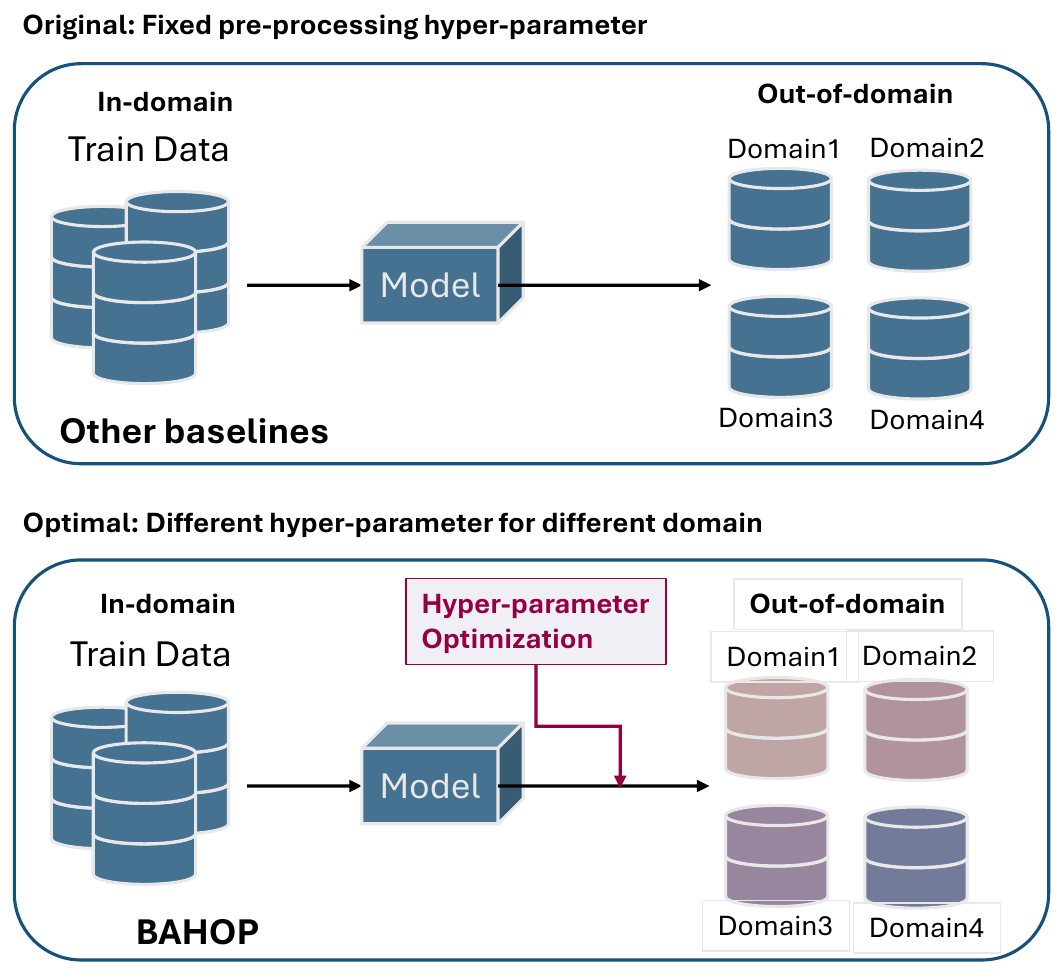}
    \caption{ The performance of out-of-domain inference varies with preprocessing parameters across various MIL models and datasets. Consequently, we suggest that each specific center within the dataset should adopt its own preprocessing parameters to maximize performance. The original method involves all centers using fixed default preprocessing hyperparameters, whereas the optimal method allows each center to use its own specific preprocessing hyperparameters determined by our proposed BAHOP.}  
    \label{fig:temp-f2}
\end{figure}
\begin{figure*}[h]
    \centering
    \includegraphics[width=0.99\textwidth]{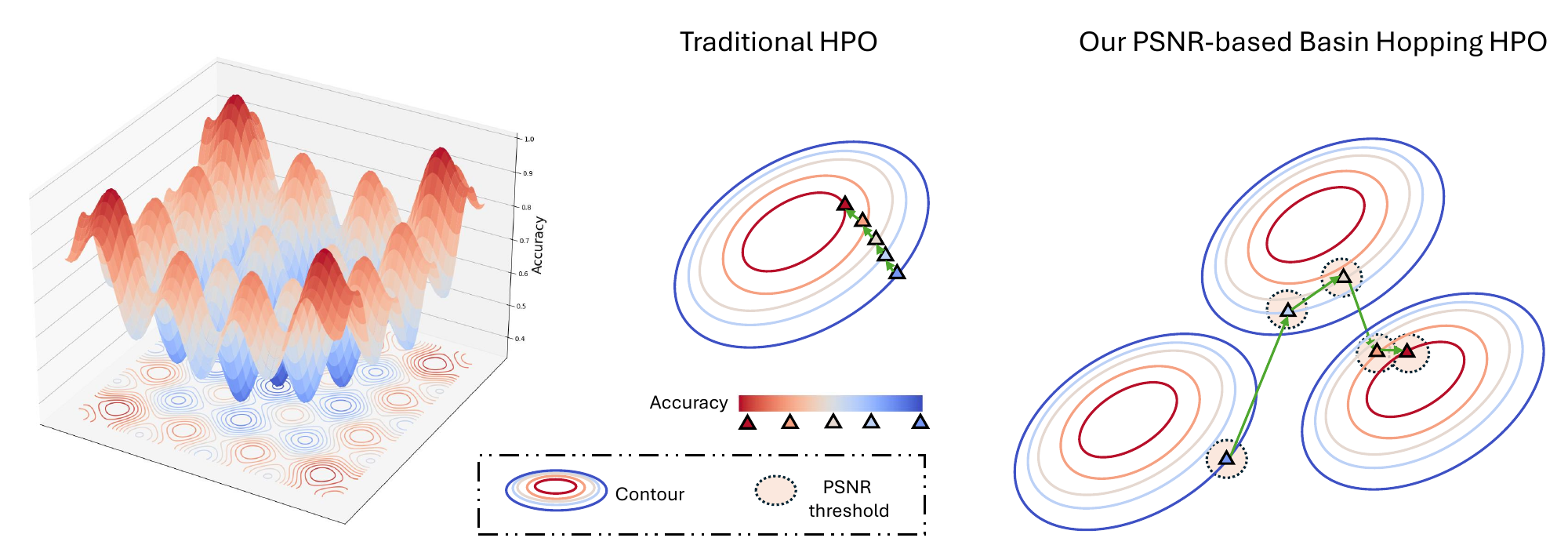}
    \caption{ PSNR-based Basin Hopping hyper-parameter optimization for out-of-domain inference in WSIs. The hyper-parameter optimization for the pre-processing task is a non-convex optimization that contains many large local optima. Our BAHOP is developed for fast search, and the jumping range of each iteration is controlled by the PSNR threshold. }  
    \label{fig:bahop}
\end{figure*}

We discovered that one factor that leads to unsatisfactory performance is the same inference hyper-parameter across different centers. 
We evaluate the fixed pre-processing parameters' performance on several sub-datasets.
%
The experiment result shows that fixed pre-processing parameters typically result in poor out-of-domain performance~\cite{howard2021impact}. For example, MIL models~\cite{lu2021data,bayesmil,dsmil} have extremely low accuracy at specific centers within the TCGA~\cite{weinstein2013cancer} and Camelyon17~\cite{c17} datasets using fixed parameters. 
Conversely, tailoring optimal pre-processing parameters for each center significantly improves out-of-domain performance (as illustrated in Figure~\ref{fig:temp-f2}).

Determining optimal pre-processing parameters for histopathological tasks is challenging.
The parameter search space is over $10^5$ due to the huge size of giga-pixel Whole Slide Images (WSIs) and the significant computational resources needed. 
%
%
Previous studies~\citep{choi2024autoft} only explore hyper-parameter optimization in learning rates and loss weight coefficients without tailoring their approaches to histopathology applications.
When adopting hyper-parameter tuning methods, traditional grid search methods for datasets like Camelyon or TCGA are inefficient, taking several hours to evaluate one parameter group. The whole hyper-parameter space contains hundreds of thousands of possible combinations. Therefore, quicker hyper-parameter search techniques are essential for histopathological tasks.

This paper focuses on how pre-processing parameters influence inference performance, particularly with out-of-domain data, and proposes an efficient parameter search method specifically designed for WSI classification. We propose a Similarity-based Basin Hopping (BAHOP) for Hyper-parameter tuning and improving the accuracy of inferring out-of-domain data across various MIL models and datasets (as shown in Fig.~\ref{fig:temp-f2}).  The key contributions of this paper are:

\begin{itemize}

\item We have observed that varying pre-processing parameters significantly impact feature extraction and, consequently, model performance—particularly in out-of-domain inference;

\item We present BAHOP, Similarity-based Basin Hopping for fast and effective parameter tuning. This algorithm enhances inference performance at Camelyon 17's center 1, boosting accuracy from 0.512 to 0.846;

\item We expand the proposed BAHOP to include other MIL models using various public datasets, achieving improvements in accuracy ranging from 5\% to 30\% for out-of-domain data across multiple MIL models and datasets;

\item The proposed BAHOP is the first fast hyper-parameter search designed explicitly for histopathological tasks.

\end{itemize}

\begin{table*}[h]
\centering
\caption{\textbf{Performance of Out-of-Domain Inference in Specific Challenging Case.} The performance is reported as the average of Accuracy, AUC, Precision, Recall, and F-score metrics, computed over the 10-fold models.}
\begin{tabular}{|l|c|c|c|c|c|c|c|c|c|c|c|c|}
\hline
\multirow{3}{*}{\textbf{Model}} & \multicolumn{2}{c|}{\textbf{In-Domain (C16)}} & \multicolumn{10}{c|}{\textbf{Out-of-Domain (C17 Center 1)}} \\
\cline{2-13}
& \multirow{2}{*}{\textbf{Acc}} & \multirow{2}{*}{\textbf{AUC}} & \multicolumn{2}{c|}{\textbf{Accuracy}} & \multicolumn{2}{c|}{\textbf{AUC}} & \multicolumn{2}{c|}{\textbf{Precision}} & \multicolumn{2}{c|}{\textbf{Recall}} & \multicolumn{2}{c|}{\textbf{F-score}} \\
\cline{4-13}
& & & \textbf{Min} & \textbf{Max} & \textbf{Min} & \textbf{Max} & \textbf{Min} & \textbf{Max} & \textbf{Min} & \textbf{Max} & \textbf{Min} & \textbf{Max} \\
\hline
\textbf{ABMIL} & 0.901 & 0.941 & 0.84 & 0.92 & 0.833 & 0.896 & 0.788 & 0.966 & 0.743 & 0.829 & 0.765 & 0.875 \\
\textbf{DSMIL} & 0.925 & 0.954 & 0.64 & 0.89 & 0.721 & 0.93 & 0.491 & 1.000 & 0.429 & 0.943 & 0.556 & 0.831 \\
\textbf{CLAM} & 0.901 & 0.946 & 0.512 & 0.847 & 0.708 & 0.833 & 0.486 & 0.839 & 0.743 & 1.000 & 0.648 & 0.788 \\
\textbf{TransMIL} & 0.892 & 0.935 & 0.73 & 0.84 & 0.861 & 0.914 & 0.730 & 0.823 & 0.753 & 0.844 & 0.724 & 0.830 \\
\textbf{Bayes-MIL} & 0.883 & 0.916 & 0.35 & 0.80 & 0.819 & 0.881 & 0.350 & 0.759 & 0.629 & 1.000 & 0.519 & 0.737 \\
\textbf{MHIM-DSMIL} & 0.925 & 0.965 & 0.77 & 0.87 & 0.850 & 0.921 & 0.607 & 1.000 & 0.600 & 0.971 & 0.710 & 0.800 \\
\hline
\end{tabular}
\label{tab:important}
\end{table*}
\section{Related Work}
\label{sec:related_work}

\paragraph{Pre-processing of WSI in histopathology.}
Variations in tissue processing techniques, including chemical fixation or freezing, dehydration, embedding, and staining, can alter the visual characteristics of the tissue slide in ways that are both non-uniform and non-linear. These changes occur across different tissue types and laboratories~\cite{asif2023unleashing,yagi2011color}, consequently, impacting the performance of deep learning.~\citet{salvi2021impact,gurcan2009histopathological,wang2024advances} summarize the pre-processing impact for WSI analysis on deep learning frameworks.~\citeauthor{tellez2019quantifying} quantifies the effects of data augmentation and stain color normalization~\cite{lee2022derivation}.

There are several research works that specifically explore the issue of dataset bias in computational pathology.~\citet{fdd} quantifies the impact of domain shift in attention-based MIL and points out that MIL performance is worse than what is reported using in-domain test data. Several works~\cite {tcga-biased,howard2021impact} discuss bias in histopathological data. Important clinical variables have been shown to be significantly different between tissue source sites, affecting the generalization issue for histopathological tasks.

\paragraph{Hyper-parameters search and optimization in machine learning.}
\cite{falkner2018bohb} propose a Bayes-optimization with Hyperband strategy and evaluate it in small datasets.~\cite{choi2024autoft} search hyper-parameters in per-layer learning rates and loss weight coefficients. These previous works are not designed for histopathology.

Hyper-parameter optimization in histopathological tasks is different than optimization problems in a linear function. The optimization in linear function or small machine learning model with a small dataset can be optimized in many iterations in a short time. However, the optimization in the machine learning model for histopathological tasks is totally different. WSIs are giga-pixel images that demand significantly higher computational resources. Despite advancements in hardware, memory constraints still pose a significant challenge in the digital processing of WSIs in current research~\cite{cifci2023ai,echle2021deep}. Additionally, nature images have very different data distribution compared to the whole slide image~\cite{raghu2019transfusion}. Therefore, designing a hyper-parameter optimization strategy for computational pathology is necessary. The proposed BAHOP algorithm is the first hyper-parameter optimization strategy designed for histopathological tasks.

\section{Motivation and Observation}
\label{motivation}

Public datasets often combine data from various domains~\cite{torralba2011unbiased,fdd}, obscuring true performance metrics within specific areas. Studies~\cite{tcga-biased,howard2021impact,hosseini2024computational} show that domain-specific variations in digital histology can affect the accuracy and bias of deep learning models. In practice, these variations necessitate frequent retraining of models due to changing data domains, which is impractical due to high computational costs. Thus, improving model performance without additional training is essential.

\subsection{Pre-processing Impacts Inference Performance for Different Domains.} 
\begin{figure}[hb]
    \centering
    \includegraphics[width=0.45\textwidth]{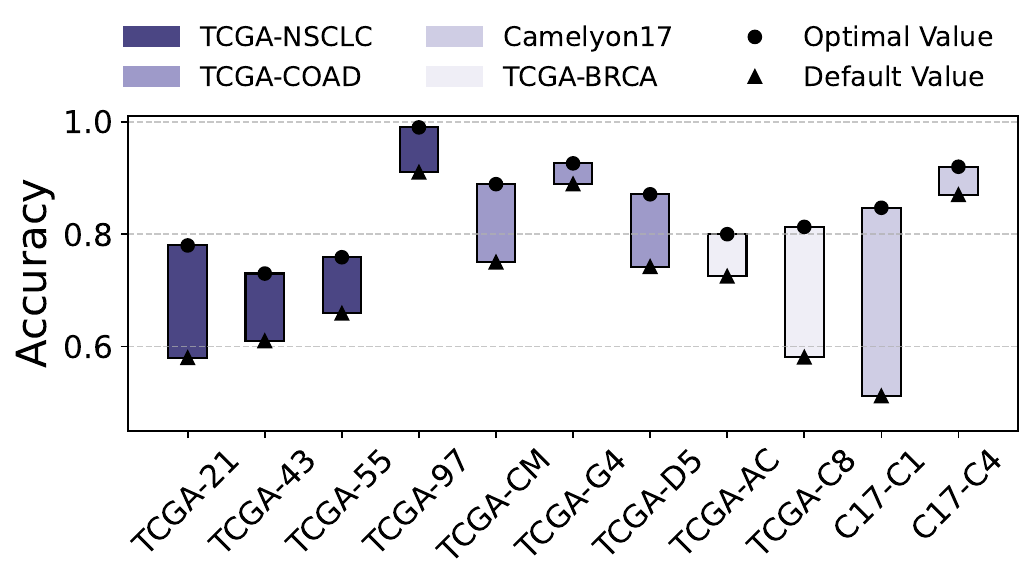}
    \caption{The inference performance of out-of-domain data varies with preprocessing parameters across various MIL models and datasets. }  
    \label{fig:f3}
\end{figure}

\begin{figure}[ht]
    \centering
    \includegraphics[width=0.47\textwidth]{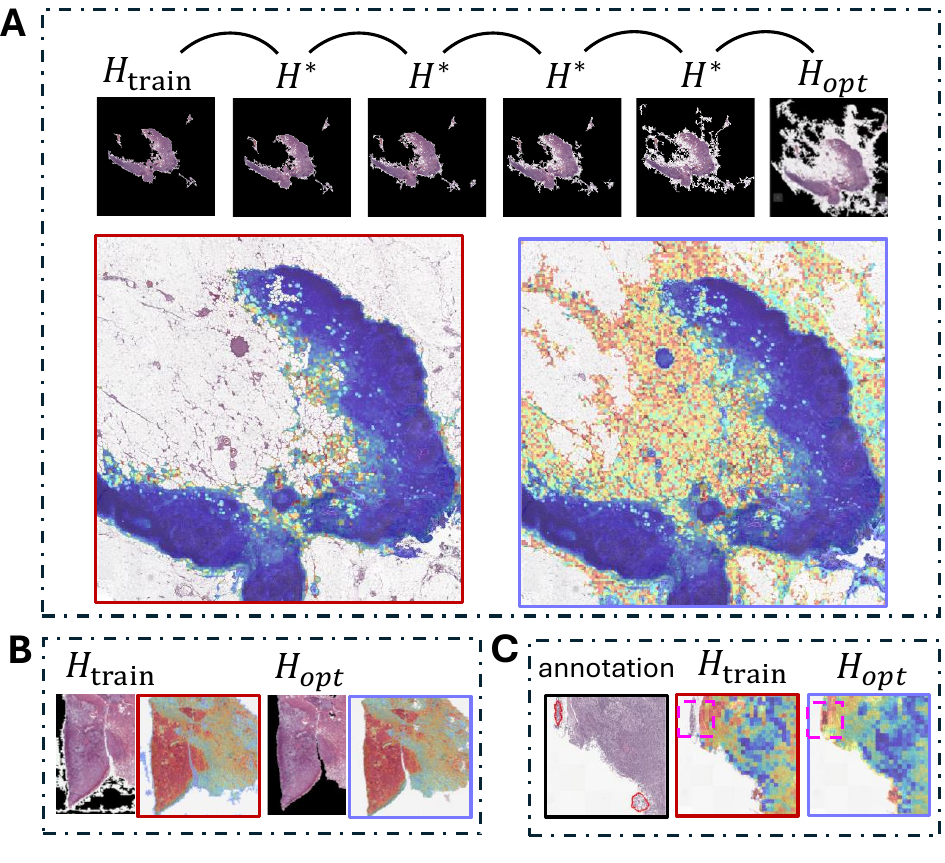}
    \caption{All the heatmaps circled by red boxes correspond to the default hyper-parameter and predict wrong, while heatmaps circled by blue boxes correspond to optimal hyperparameters with correct predictions. Fig.A: Hyper-parameter optimization starts from the default parameter, which is the same as the pre-trained model. The default pre-processed WSI drops many patches that get a high attention score in optimal pre-processed WSI. Fig.B, the dataset is TCGA-NSCLC where dropping tissue can improve accuracy. Fig.C: Some hyper-parameters drop the region of interest(ROI) during pre-processing.}  
    \label{fig:psnr_attention}
\end{figure}

We found there exists significant variability in inference performance on out-of-domain data across various methods (refer to Table~\ref{tab:important}). We employ a grid search to fully understand how pre-processing parameters influence outcomes in center 1 of Camelyon 17 in this motivation experiment. For instance, the CLAM~\cite{lu2021data} model trained on the Camelyon16 dataset achieves an overall inference accuracy of 0.86 on the entire Camelyon17 dataset but drops to 0.512 at Center 1 of Camelyon17, indicating inconsistent performance. We also evaluate our approach in other centers across different datasets, as detailed in Fig~\ref{fig:f3} and Appendix. Similar variations are observed across other datasets and MIL models, and the out-of-domain inference is extremely poor in some cases. 
What's more, the default pre-processing setting yields varying results depending on the center as shown in Fig~\ref{fig:f3}, where identical hyper-parameters may perform well in one center but poorly in another.

\paragraph{Observation 1} The effection of the pre-processing parameters varies across different datasets.
%
As shown in Fig.~\ref{fig:psnr_attention} and the Appendix, the detection of foreground contours, controlled by several hyperparameters, often results in the exclusion of large areas, which are treated as holes. For instance, adipocyte cells are frequently excluded. However, some tumor cells surrounded by fibers, adipocyte cells, or lipid droplets may also be filtered out due to preprocessing parameters and discarded. In Fig.~\ref{fig:psnr_attention}(a) and (c), excessive patch removal leads to incorrect predictions, while in Fig.~\ref{fig:psnr_attention}(b), it improves prediction accuracy when applied to a different dataset. This variability makes it difficult to quantify the impact of specific patches, especially since the number of patches for a single WSI in a MIL model can exceed \(10^5\). While some studies, such as \citet{javed2022additive}, address this issue, there is no consensus on how specific patches influence machine learning models in computational pathology. An adaptive algorithm is needed to address this challenge in histopathological tasks.

\paragraph{Observation 2} 
Pre-processing optimization is inherently a non-convex problem with numerous large local optima (as shown in Fig.~\ref{fig:bahop}). Our objective is to achieve an optimal result with reduced time and computational costs, where this optimum does not need to be a global one.

Rather than identifying specific cell types within tissue areas, we approach this histopathological task from a machine learning perspective. The pre-processing of WSIs focuses on segmenting tissue regions and discarding patches identified as holes or backgrounds. These patches are selected and filtered based on pre-defined parameters, while most tissue areas containing cells are retained. Consequently, the variations in inference performance caused by multiple interacting hyperparameters can be framed as a hyperparameter optimization problem. This problem can be effectively addressed using an adaptive algorithm tailored for this histopathological task.

\subsection{Inefficient Manual Pre-Processing Parameter Search.}

In real-world scenarios, retraining machine learning models for every new dataset is often impractical. While optimizing parameter performance is advantageous, finding the optimal settings is highly time-consuming. For instance, evaluating a single set of parameters, including feature extraction, takes approximately 7 hours on an RTX3080 (details in Appendix). Traditional optimization techniques, such as tuning learning rates, batch sizes, or loss weight coefficients, cannot be directly applied to histopathological tasks. Consequently, more efficient parameter search methods are required. Common strategies like grid search demand excessive computational resources and time, rendering them unsuitable for histopathological applications.

Based on these observations, we want to explore three questions.
\begin{itemize}
    \item Can we improve inference performance on out-of-domain data just by modifying the preprocessing parameters?
    \item How can we search for parameters more cost-effectively to enhance inference performance?
    \item Do these observations apply to other multiple instance learning (MIL) models?
\end{itemize}

Therefore, we introduce our approach in a fast hyper-parameter search in pre-processing parameters of histopathology.

\section{Fast Hyper-Parameter Search for Inference Performance in Out-Of-Domain Data }
In this section, we formalize hyperparameter optimization and present our BAHOP algorithm step by step. The proposed BAHOP algorithm includes PSNR-based mechanism for limiting the current searching space in each iteration and Basin Hopping to avoid local optima.

\subsection{Definition of Problem}
A good choice of hyperparameters can significantly improve out-of-domain inference performance, yet optimal hyperparameters vary, depending on the datasets and the models. 
We assume access to two types of datasets: (1) a large dataset $D_{train}$ for training to produce the foundation model subsequently used to test out-of-domain performance. (2) Multiple subdataset $\{D_{val} = D_{val}^1, D_{val}^2, \cdots, D_{val}^n\}$. The validation sub-dataset $D_{val}$ is much smaller than $D_{train}$ and is used only for optimization at the outer level, not for fine-tuning. Note that these test distributions are distinct and are never shown to the model before the final evaluation.

Let $M$ denote a pre-trained foundation model trained in a large dataset $D_{train}$ with the pre-processing parameter $H_{train}$. Now, there are many out-of-domain datasets $\{D_{val} = D_{val}^1, D_{val}^2, \cdots, D_{val}^n\}$. They are pre-processed by hyper-parameter $H_1, H_2, \cdots, H_n$ to obtain the features $F_{val}^1, F_{val}^2, \cdots, F_{val}^n$ for inferring. The out-of-domain performance of model $M$ would be tested by the features $F_{val}^n$ that are preprocessed by the hyper-parameter $H_n$ as $ y = Infer(H_n, D_{val}^n)$, where $y$ is the performance metric.

We define the histopathological task of identifying the optimal pre-processing for enhanced out-of-domain performance as a hyper-parameter optimization problem, as described in Equation~\ref{form}, where $\Phi$ is the hyper-parameter space. This process helps us determine the most effective final hyper-parameters $ H_n^*$ for each validation center $D_{val}^n$.  

\vspace{-0.5cm}
\begin{equation}
 H_n^* = \underset{H \in \Phi}{\operatorname{arg\,max}} \, \text{Infer}(BAHOP(H, D_{val}^n), D_{val}^n)
\label{form}
\end{equation}

\subsection{Pre-Processing Problem Modeling}
Next, we consider the optimization problem for histopathological tasks. The optimization of $H_n$ is actually the optimization of a combination of pre-processing parameters, not a single variable. 
Previous pre-processing of WSIs encompasses a series of standard procedures. Initially, tissue segmentation is executed automatically on each slide at a reduced magnification. This involves generating a binary mask for the tissue regions by applying a binary threshold to the saturation channel of the downsampled image, following its conversion from RGB to HSV color space. Morphological closing and median blurring are employed for smoothing the contours of the detected tissue and minimizing artifacts. Subsequently, the approximate contours of both the tissue and the tissue cavities are assessed and filtered based on their area, culminating in the creation of the final segmentation mask. 

We complete set of pre-processing hyperparameters for optimization in histopathological tasks is thus $H_n = [x_1, x_2,x_3,x_4,x_5,x_6]$, where each $x_i$ has its own hyper-parameter space. As illustrated in Fig~\ref{fig:psnr_attention} and the observations discussed in Section~\ref{motivation}, the set of pre-processing hyperparameters affect the performance together, varying across the different datasets. A large area of tissue is detected as background or holes and then dropped during the pre-processing. As illustrated in Fig~\ref{fig:psnr_attention}, dropping too many patches decreases performance in Fig.~\ref{fig:psnr_attention}(a) and (c) but increases performance in Fig.~\ref{fig:psnr_attention}(b) where the dataset is changed. It is hard and no work to measure how these patches and specific types of cells affect the out-of-domain inference performance in histopathological tasks. 
Thus, we optimize these histopathological tasks by machine learning techniques instead of exploring the impact of the specific type of cells.

As illustrated in Fig~\ref{fig:bahop}, the hyper-parameter optimization for histopathological tasks is non-convex optimization which has many large local optima. Therefore, our goal is to search for an optimal hyper-parameter in a short time and with less cost; the optima we search for can just be larger as it can.  We are not searching for the global optima. We just search for an optimal value much better than the default hyper-parameter in a very short time and cost. In practice, this involves repeatedly applying different features within the same foundational model using various hyperparameters $H_n^*$. Therefore, we propose a PSNR mechanism as a threshold to advance the speed of hyper-parameter search.

\subsection{Optimization of Fast Hyper-Parameter Search}

\subsubsection{Good Pre-Processed WSIs show High Structural Similarity.}
\begin{figure}[hb]
    \centering
    \includegraphics[width=0.4\textwidth]{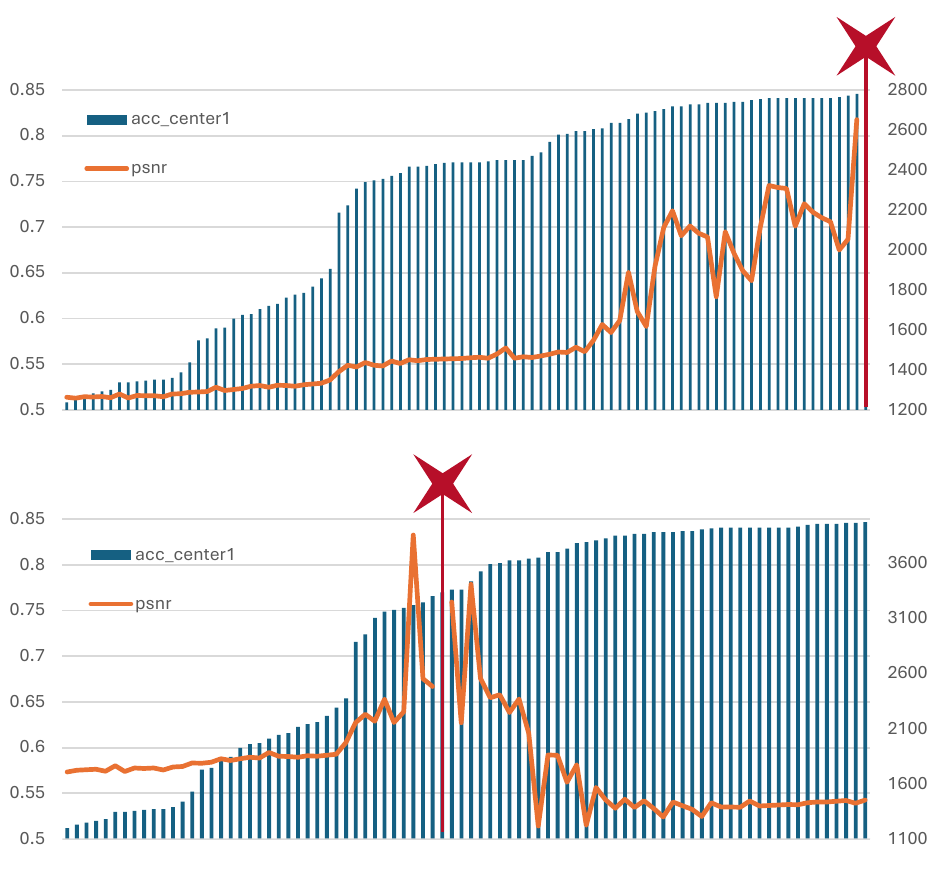}
    \caption{ Relationship between PSNR and Accuracy. The red star stands for the reference object. All the hyper-parameters are compared with the reference object to calculate the PSNR.  }  
    \label{fig:psnr_acc}
\end{figure}

For single-domain inference performance, our experiments demonstrate that pre-processing parameters, which lead to similar high accuracy, also generate the pre-processed images with similar structures(as illustrated in Figure~\ref{fig:psnr_acc}). When the two objects are close in accuracy, their PSNR is always high. This phenomenon always exists, no matter how we change the reference object.

Therefore, we only evaluate the pre-processed Whole Slide Images (WSIs) that closely resemble the current optimally pre-processed WSI in each optimization iteration. To quantitatively assess differences between two sets of pre-processing results, we use Peak Signal-to-Noise Ratio (PSNR) to measure the similarity of images generated during patch creation. The general histopathological workflow includes creating patches, extracting features, training models, and testing inference data. During patch generation, thumbnails are created for visualizing the preprocessing effects. Thumbnails are 64x downsampled from original giga-pixel WSIs. Thus, we can determine the similarity between these two sets of pre-processed results by calculating PSNR for these thumbnails efficiently.

\subsubsection{Peak Signal-to-Noise Ratio (PSNR) }

Peak Signal-to-Noise Ratio (PSNR) is a common measurement in image processing and digital signal processing. A higher PSNR value indicates greater similarity between two images. We calculate the PSNR by comparing thumbnails generated during preprocessing. For example, the total PSNR of thumbnails produced using the second-best preprocessing parameters should be higher than that of any other set, as illustrated in Figure~\ref{fig:psnr_acc}.

\subsubsection{PSNR-based Basin Hopping for Hyper-Parameter Optimization}

\begin{algorithm}[h]
\caption{PSNR-based Basin Hopping Hyper-Parameter Optimization \textbf{(BAHOP)} }
\label{alg:sa_patch_similarity}
\textbf{Input}: Default parameter ${H_{train}}$ with result set $\mathcal{R}$, Hyper-parameters space $\Phi$, initial similarity threshold $\tau$, function $Infer$ to evaluate performance, function $S$ to calculate similarity\\
\textbf{Output}: Optimal parameters $H^*$
\begin{algorithmic}[1]
\STATE Initialize $H^*$ with fixed default parameter ${H_{train}}$
\STATE Generate pre-processed thumbnails $I^*$ from $H^*$ (for all validation WSIs)
\STATE Generate feature $F^*$ from WSIs with $H^*$
\STATE $y^* \gets Infer(F^*)$
\STATE $\mathcal{R} \gets  (H^*,y^*) $
\WHILE{optimization times}
    \STATE Select $H_{\text{new}}$ by perturbing $H^*$ slightly from $\Phi$
    \IF{ $H_{\text{new}}$ not in $\mathcal{R}$} 
        \STATE Generate thumbnails $I_{\text{new}}$ with $H_{\text{new}}$
        \IF{$S(I_{\text{new}}, I^*) > \tau$ }
            \STATE Extract feature $F_{\text{new}}$ 
            \;\;\;\; $\triangleright$ Expensive step
            \STATE $y_{\text{new}} \gets Infer(F_{\text{new}})$  
            \STATE $\mathcal{R} \gets \mathcal{R} \cup (H_{\text{new}}, y_{\text{new}})$
            \IF{$y_{\text{new}} > y^*$  }
                \STATE $H^*, I^*, y^* \gets {H_\text{new}}, I_{\text{new}}, y_{\text{new}}  $ 
            \ENDIF
        \ENDIF    
    \ENDIF

\ENDWHILE
\RETURN $H^*$
\end{algorithmic}
\end{algorithm}

Basin-hopping is a global optimization technique characterized by iterative processes involving random perturbation of coordinates, followed by localized optimization and subsequent evaluation and acceptance or rejection of new coordinates based on minimized function values. This method is particularly useful in high-dimensional landscapes. However, an inherent limitation of Basin-hopping lies in the necessity to execute random perturbations at a designated point, typically a local minimum. These perturbations must be suitably sized - adequately large to escape the current local minimum yet not excessively to devolve into total randomness. To address this challenge, we introduce a Peak Signal-to-Noise Ratio (PSNR) threshold within the Basin-hopping optimization process. Subsequently, our experimental endeavors focus on establishing a correlation between pre-processing methodologies applied to whole slide images and the PSNR metric.

The integration of PSNR-based filtering does not fundamentally change the Basin Hopping process itself but rather serves as a heuristic pre-processing step to enhance efficiency and fit histopathological tasks. The PSNR threshold is set dynamically and adapts across different datasets. The experiments of the effection of each pre-processing parameter are shown in the Appendix. From the experiments (as shown in the Appendix), the criteria for selecting the PSNR threshold and dynamically adapting to different datasets is that we perturb the pre-processing parameter -- segmentation threshold (related to foreground and background detection). Each time, we change the value to plus or minus 1. We calculate the PSNR for this comparison as the initial PSNR. This process only requires the creation of a patch and quick PSNR calculation in thumbnails.

Also, we delete the mechanism of temperature and accepted probability from the original Basin hopping, ensuring our BAHOP is designed properly for histopathological tasks. For our proposed BAHOP method (as shown in Algorithm~\ref{alg:sa_patch_similarity}), we use ${H_{train}}$ as the default parameter. Using the current optimal parameter $H^*$, we generate a new parameter ${H_{new}}$, create patches, and calculate their PSNR against the optimal thumbnails $I^*$. If the PSNR exceeds threshold $\tau$, we proceed to more resource-intensive tasks such as feature extraction in histopathological analysis and subsequent inference using an existing model. Each new parameter set ${H_{new}}$ is added to our historical dataset $R= \{(H_1,y_1),...,(H_i,y_i)\}$. If ${H_{new}}$ demonstrates improved performance, ${H_*}$ will be updated with ${H_{new}}$. 

\section{Experiments}

\begin{table*}[th]
\centering
\begin{tabular}{l|l|l|l|cc|l|l|l|cc}
\hline
\multirow{3}{*}{Model} & \multicolumn{5}{c|}{TCGA-COAD}  & \multicolumn{5}{c}{TCGA-BRCA}  \\
\cline{2-11}
& \multirow{2}{*}{Center}  & \multicolumn{2}{c|}{Accuracy} & \multicolumn{2}{c|}{AUC}& \multirow{2}{*}{Center}  & \multicolumn{2}{c|}{Accuracy} & \multicolumn{2}{c}{AUC}  \\

& &  Default & Ours & Default & Ours &  & Default & Ours & Default & Ours \\
\hline
\multirow{4}{*}{ABMIL~\citep{abmil}} 
&CM & 0.75 & 0.778 & 0.828&0.838  &  A7 & 0.902&0.961 &0.964&0.994 \\
&D5 & 0.709 & 0.774 & 0.64&0.647  &AC & 0.775&0.805 &0.728&0.761 \\
& DM & 0.783 & 0.826 &  0.732&0.759&AR & 0.781 &0.816 & 0.834& 0.856\\
& G4 & 0.963 & 0.963& 0.98&1  & C8 & 0.907&0.93 & 0.867 &0.908\\
\hline
\multirow{4}{*}{DSMIL~\citep{dsmil}} 
& CM & 0.75  & 0.778 & 0.848&0.859 &  A7 & 0.921&0.941 &0.923&0.962 \\
& D5 &  0.806 & 0.871&0.5 &0.493  &AC &  0.725&0.805 &0.761 &0.761\\
& DM & 0.869  & 0.869 &0.777 &0.804&AR & 0.765&0.781& 0.733& 0.738 \\
& G4 &  0.963 & 0.963 &0.96& 0.98 & C8 & 0.581&0.813&0.767& 0.75 \\
\hline
\multirow{4}{*}{CLAM~\citep{lu2021data}} 
& CM & 0.75 &  0.889 & 0.56 &0.636  & A7 & 0.941&0.961  &0.947 &0.974 \\
& D5 & 0.742  &0.871  & 0.46& 0.793 &AC &  0.75&0.81 &0.772 &0.761  \\
& DM & 0.739  & 0.739 & 0.571 & 0.571 &AR &  0.813&0.828 &0.882 &0.878 \\
& G4 &  0.889 & 0.926  & 0.94  &0.96 & C8 &  0.907&0.93 & 0.867 & 0.875 \\
\hline
\multirow{4}{*}{BayesMIL~\citep{bayesmil}} 
& CM &  0.861&0.889 & 0.929 & 0.929  & A7 & 0.882&0.882 &0.934 & 0.955  \\
& D5 &  0.709 & 0.806  &0.667 & 0.593 &AC & 0.775&0.775  & 0.799 &0.821 \\
& DM &  0.783 & 0.826  & 0.821 &0.875 &AR & 0.813&0.828  & 0.796 & 0.791 \\
& G4 &  0.926  & 1 & 0.96 &1  & C8 & 0.86&0.884   & 0.792 & 0.767 \\
\hline
\multirow{4}{*}{\shortstack{MHIM-\\
DSMIL}~\citep{mhim-mil} }
& CM & 0.75  & 0.801 & 0.77& 0.798 &  A7 &  0.923&0.961 & 0.957 & 0.996  \\
& D5 &  0.839 & 0.871 &0.48& 0.493 &AC &  0.75&0.775&0.611 & 0.745 \\
& DM &  0.826 & 0.87 & 0.714& 0.786 &AR & 0.75&0.766&0.709& 0.713 \\
& G4 &  0.925 & 0.963 &0.94&0.98 &C8 & 0.907&0.93 &0.742&0.733 \\
\hline
\end{tabular} 

\caption{\textbf{Experiments about out-of-domain inference performance in TCGA-COAD and TCGA-BRCA for cancer subtype.} We compare the accuracy obtained by default hyper-parameters with the optimal hyper-parameter searched by our BAHOP algorithm. }
\label{results}
\end{table*}

\begin{table*}[h]
\centering
\begin{tabular}{|l|c|c|c|c|c|c|c|c|c|c|c|c|c|c|}
\hline
\multirow{3}{*}{\textbf{Model}}  & \multicolumn{4}{c|}{Out-of-Domain(Center 0)} & \multicolumn{4}{c|}{Out-of-Domain(Center 1)} &\multicolumn{4}{c|}{Out-of-Domain(Center 4)} \\
\cline{2-13}
& \multicolumn{2}{c|}{Accuracy} & \multicolumn{2}{c|}{AUC}  & \multicolumn{2}{c|}{Accuracy} & \multicolumn{2}{c|}{AUC} & \multicolumn{2}{c|}{Accuracy} & \multicolumn{2}{c|}{AUC} \\

&  Def & Ours & Def & Ours &  Def & Ours & Def & Ours &  Def & Ours & Def & Ours  \\
\hline
ABMIL~\cite{abmil}  & 0.93&0.97& 0.921& 0.965 &0.84&0.92& 0.833&0.896 & 0.88&0.93  &0.907 & 0.911\\
DSMIL~\cite{dsmil} & 0.77 &0.96 &0.88&0.937& 0.64&0.89& 0.721&0.93&0.92&0.96&0.952 & 0.947\\
CLAM~\cite{lu2021data} & 0.92 &0.95 &0.919 &0.963 &0.512 &0.846 &0.708&0.833&0.874&0.92  &0.904 &0.941\\
Bayes-MIL~\cite{bayesmil} & 0.83 & 0.94  & 0.956&0.884&0.35& 0.80&0.819&0.881 &0.85&0.92 & 0.931&0.899 \\
MHIM-DSMIL~\cite{mhim-mil} & 0.87 &0.96 &0.911 &0.956 & 0.77&0.87&0.85&0.921&0.91 &0.94  &0.949 & 0.955\\
\hline

\end{tabular}
\caption{Out-of-domain inference performance in \textbf{Center 0, center1 and Center 4 of Camelyon17} for normal and tumor classification.}
\label{c17_performance}
\end{table*}

\subsection{Experiments Setup}


Our experiment utilizes the pre-processing pipeline from CLAM~\cite{lu2021data}. The datasets include Camelyon 16, Camelyon 17~\cite{c17}, TCGA-NSCLC, TCGA-BRCA, and TCGA-COAD~\cite{weinstein2013cancer}. For Camelyon16 and Camelyon17, we focus on normal-tumor classification; for all TCGA datasets, we address cancer subtype tasks. In our experiments, we fix the seeds and model hyperparameters such as learning rate and loss weight across similar tasks within the same model framework. Additional experimental details are provided in the Appendix. We employ K-fold, Stratified K-Fold or K-fold Monte Carlo cross-validation methods (with k sets to 10) to train models and assess out-of-domain performance. All results presented in Table~\ref{results}, Table~\ref{c17_performance} and Table~\ref{tab:nsclc} represent average accuracy calculated over 10-fold models.

\subsection{Classification Accuracy Validation of BAHOP}
For the Camelyon datasets, we use 270 WSIs from Camelyon 16 to train the model and test inference performance across various centers in Camelyon17 to simulate out-of-domain scenarios. It's important to note that Camelyon16 data was collected from two centers (UMCU and RUMC), while Camelyon17 includes data from five centers (CWZ, RST, UMCU, RUMC, and LPON). Consequently, only centers 0 (CWZ), center 1 (RST), and center 4 (LPON) of Camelyon17 represent out-of-domain situations.

We compare inference performance using features extracted by default parameters (not the worst) and those optimized through our BAHOP algorithm. As shown in Table~\ref{results} and Table~\ref{c17_performance}, the optimal hyper-parameters identified by our BAHOP algorithm generally surpass the default settings across various models and datasets. In specific cases, such as CLAM and BayesMIL at Center 1 of Camelyon 17, performance improved dramatically from 0.512 to 0.847 simply by adjusting the hyper-parameters.

For TCGA datasets, we divide the entire dataset according to TCGA Tissue Source Site Codes~\cite{weinstein2013cancer}. We select several centers with sufficient WSIs to form the testing sub-dataset, while the remaining data constitutes the training dataset. The out-of-domain inference performance also improves (as shown in Table~\ref{results}), indicating that our issue is general and our BAHOP provides a universal solution. Additionally, we conduct experiments on a larger TCGA-NSCLC dataset that includes more extensive out-of-domain testing (as detailed in Table~\ref{tab:nsclc}).

\begin{table}[h]
\begin{tabular}{l|c|c|c|c|c|c}
\hline
\multicolumn{7}{c}{TCGA-NSCLC} \\
\hline
 \multirow{2}{*}{Center} & \multicolumn{2}{c|}{ABMIL~\cite{abmil}} & \multicolumn{2}{c|}{CLAM~\cite{lu2021data}} & \multicolumn{2}{c|}{Bayes-MIL~\cite{bayesmil}} \\ 
\cline{2-7}
 & Def.& Ours &Def. & Ours & Def. & Ours \\ 
\hline

 21 & 0.72 & \textbf{0.78} & 0.56 & \textbf{0.78} & 0.56 & \textbf{0.67} \\
 22 & 0.87 & \textbf{0.89} & 0.78 &\textbf{ 0.84 }& 0.84 & 0.84 \\
 43 & 0.75 & \textbf{0.80} & 0.60 & \textbf{0.75} & 0.70 & \textbf{0.80} \\
 49 & 0.59 & \textbf{0.64} & 0.69 & \textbf{0.71} & 0.71 & \textbf{0.76} \\
 55 & 0.49 & \textbf{0.52} & 0.66 & \textbf{0.76} & 0.28 & 0.33 \\
 77 & 0.93 & 0.93 & 0.91 & \textbf{0.93} & 0.75 & \textbf{0.80} \\
 97 & 1.00& 1.00 & 0.91 & \textbf{1.00} & 0.87 & \textbf{0.96} \\
\hline
\end{tabular}
\caption{ Comparison of inference accuracy in TCGA-NSCLC with more centers. Only accuracy is reported since each separate center only has one category. Def. stands for default parameter.}

\label{tab:nsclc}
\end{table}

\subsection{Comparison of other hyper-parameter optimization}

As illustrated in Table~\ref{tab:Basin}, we compare different hyper-parameter optimization (HPO) techniques in one center of Camelyon 17 over 100 iterations. Our BAHOP can find the optimal hyper-parameters much quicker compared to alternative HPO methodologies. While other HPO techniques demonstrated the capability to achieve commendable accuracy levels, they incurred substantial costs in terms of computational time and resource utilization.

\begin{table}[ht]
    \centering
    \caption{Comparison of other hyper-parameter optimization. SA stands for Simulated Annealing. Lat. stands for Latency (smaller is better) and Mem.
stands for Memory. }
    \begin{tabular}{l|r|r|r}
    \hline
    \textbf{Strategy} & \textbf{Acc} & \textbf{Lat.(min)} & \textbf{Mem.(GB)}\\
    \hline
    Random search& 0.845 & 9600 & 1250\\
    Grid search &0.847&9600 &  1250 \\
    SA~\citep{van1987simulated}  & 0.846 & 9600&  1250 \\
    Bayes OPT~\citep{bayesopt} &  0.834 &9600 & 1250 \\
    \textbf{BAHOP (Ours)}& \textbf{0.846} & \textbf{1770} &\textbf{170} \\
    \hline
    \end{tabular}
    \label{tab:Basin}
\end{table}

\subsection{Efficiency Validation}

\begin{figure}[h]
    \centering
    \includegraphics[width=0.49\textwidth]{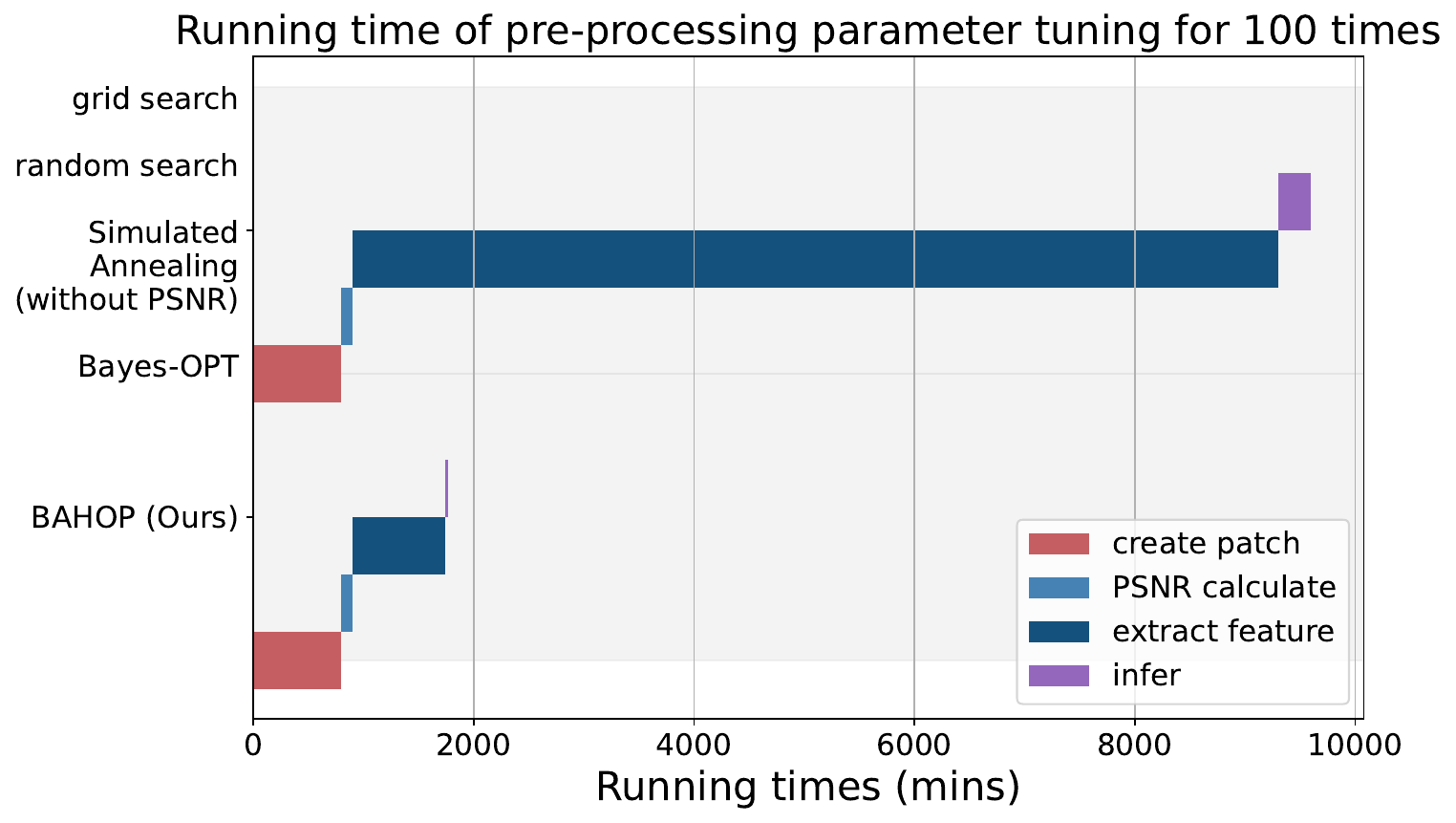}
    \caption{\textbf{The running time} of pre-processing parameters tuning for 100 times. Our algorithm can save time by skipping a large number of feature extraction. }
    \label{fig:running}
\end{figure}

Figure \ref{fig:running} shows the running time for optimizing preprocessing parameters 100 times. Our PSNR-based Basin Hopping (BAHOP) algorithm for Parameter Tuning uses PSNR to compare each pre-processed WSIs with the best one so far, selecting only those with high similarity for further feature extraction. WSIs with low PSNR are skipped, streamlining the process and ensuring that only promising candidates are evaluated further. This approach prevents our BAHOP from getting stuck in local optima and saves time by reducing unnecessary feature extractions.

\subsection{Ablation Study}

\paragraph{Strategy 1: PSNR} 

The workflow of inference in histopathological tasks includes creating patches, extracting features, and inferred in a pre-trained model. Without the PSNR-based strategy, the algorithm will run the most expensive step -- feature extraction, for each iteration. Our PSNR machanism can skip feature extraction in many iterations (as shown in Table~\ref{tab:ab}).

\paragraph{Strategy 2: Basin Hopping}

Basin Hopping prevents optimization from getting stuck in local optima. As illustrated in Table~\ref{tab:Basin}, search algorithms lacking Basin Hopping repeatedly test hyper-parameters that yield a higher PSNR than the current optimal solution. Since other optimizaion techniques have optimized for 100 times, the optimal accuracy is all high but the costs are so huge both in time and disk space.

\begin{table}[h]
    \centering
    \caption{\textbf{ Ablation Study of the PSNR and Basin Hopping.} Results are based on optimizing 100 times in the center1 of Camelyon 17. Lat. stands for Latency (smaller is better) and Mem. stands for Memory. }
    \begin{tabular}{l|c|c|c|c}
    \hline
      \textbf{PSNR}& \textbf{BH} &  \textbf{Acc}& \textbf{Lat. (min)}&\textbf{Mem. (GB)} \\

    \hline
    \XSolidBrush &  \CheckmarkBold  &0.846 & 9600 & 1250 \\
   \CheckmarkBold   & \XSolidBrush  & 0.845 &2553  &278\\
     \CheckmarkBold &  \CheckmarkBold & \textbf{0.846} & \textbf{1770} & \textbf{170} \\
    \hline
    \end{tabular}
    \label{tab:ab}
\end{table}

\section{Conclusion}

Different pre-processing parameters significantly impact feature extraction and model performance in histopathological tasks. In this paper, we propose the Similarity-based Basin Hopping (BAHOP) algorithm for fast parameter tuning, which enhances inference performance on out-of-domain data. BAHOP achieves a 5\% to 30\% accuracy improvement on the Camelyon and multiple TCGA datasets, offering faster hyper-parameter search by reducing feature extraction steps based on WSI characteristics.

{
    \small
    \bibliographystyle{ieeenat_fullname}
    \bibliography{main}
}


\end{document}